\documentclass{bmvc2k}

\usepackage{multirow}
\usepackage{pbox}
\usepackage{enumitem}
\usepackage{amsmath}
\usepackage{bbm}

\title{Equal But Not The Same: Understanding the Implicit Relationship Between Persuasive Images and Text}

\addauthor{Mingda Zhang}{mzhang@cs.pitt.edu}{1}
\addauthor{Rebecca Hwa}{hwa@cs.pitt.edu}{1}
\addauthor{Adriana Kovashka}{kovashka@cs.pitt.edu}{1}

\addinstitution{
 Department of Computer Science\\
 University of Pittsburgh\\
 Pittsburgh, PA, USA
}

\runninghead{Zhang, Hwa, Kovashka}{
Equal But Not The Same
}

\begin{document}

\maketitle

\vspace{-1em}
\begin{abstract}
Images and text in advertisements interact in complex, non-literal ways. The two channels are usually complementary, with each channel telling a different part of the story. 
Current approaches, such as image captioning methods,
only examine literal, redundant relationships, where image and text show exactly the same content. To understand more complex relationships, we first collect a dataset of advertisement interpretations for whether the image and slogan in the same visual advertisement form a \textit{parallel} (conveying the same message without literally saying the same thing) or \textit{non-parallel}relationship, with the help of workers recruited on Amazon Mechanical Turk. 
We develop a variety of features that capture the creativity of images and the specificity or ambiguity of text, as well as methods that analyze the semantics within and across channels. We show that our method outperforms standard image-text alignment approaches on predicting the parallel/non-parallel relationship between image and text.
\end{abstract}

\vspace{-1em}
\section{Introduction}
\label{sec:intro}
\vspace{-0.5em}

Modern media are predominantly multimodal. Whether it is a news article, a magazine story, or an advertisement billboard, it contains both words and images (and sometimes audios and videos as well). Effective communications require the different channels to  
complement each other in interesting ways. For example, a mostly factual news report about a politician might nonetheless be accompanied by a flattering photo, thus subtly   
conveying some positive sentiments towards the politician. A car ad might juxtapose evocative adjectives such as ``powerful'' next to images of horses and waterfalls, inviting the viewers to make the metaphorical connections.

While recent work has made advances in making \textit{literal} connections between image and text (e.g., image captioning, where the text describes what is seen in the image \cite{vinyals2015show,Karpathy_2015_CVPR,Donahue_2015_CVPR,Johnson_2016_CVPR,Hendricks_2016_CVPR,Vedantam_2017_CVPR,show_adapt_tell,Shetty_2017_ICCV}), recognizing \textit{implicit} relationships between image and text (e.g., metaphorical, symbolic, explanatory, ironic, etc.) remains a research challenge.  


In this work, we take a first step toward the automatic analysis of non-literal relations between visual and textual persuasion.  
To make the problem concrete, we focus on the relationship between the visual component of an image advertisement and the slogan embedded in the ad. 
In particular, we 
propose a method for determining whether the image and the slogan are in a \textit{parallel} relationship, i.e. whether they convey the same message independently. 
%

\begin{figure}[t]
\includegraphics[width=1\linewidth]{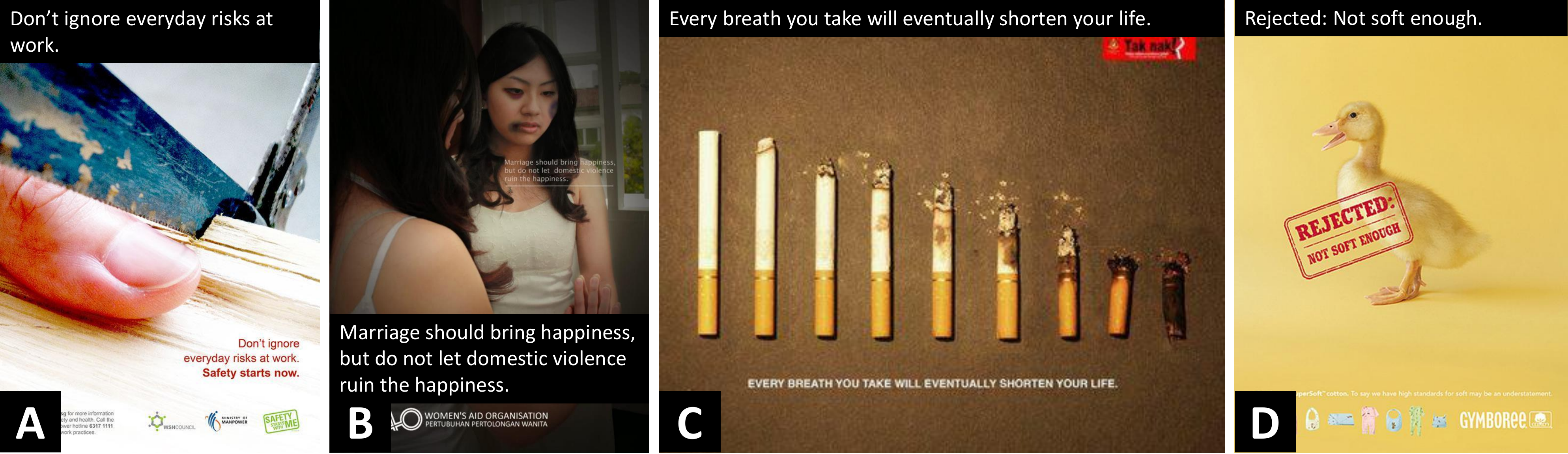}
\caption{Images and text in persuasive advertisements reinforce each other in complex ways. Even for image-text pairs that are \textit{parallel}, the image and text are complementary rather than redundant (e.g. the first image shows a saw but the text only mentions risks). This is in contrast to image captioning approaches that assume redundant image-text pairs.}
\vspace{-1em}
\label{fig:concept}
\end{figure}

Cases A and B in Fig.~\ref{fig:concept} are two examples of the \textit{parallel} relationship:
the image and text in A each warns of commonplace dangers; 
the image and text in B each warns of unhappiness caused by abuse.  
Note, however, that in both cases, the image and text do not obviously repeat the same concept:
the text in A does not mention any finger or a saw, and the image in B only subtly suggests the aftermath of violence. These examples are challenging because their messages are parallel \textit{but not equivalent}.
Cases C and D are examples of the \textit{non-parallel} relationship. In C, the text alone does not establish the topic of smoking, and the image alone does not establish a  connection between the shortening of the cigarettes' lengths and the shortening of one's life. The image and text in D even appear to be contradictory, because a fuzzy duckling is ``not soft enough.''

As the above examples show, elements from the image and text do not always align, suggesting that standard image-text alignment methods alone are insufficient for detecting parallel relationships.
We hypothesize that additional cues, such as surprise, ambiguity, specificity and memorability, are needed. To validate our claim, we first collected a dataset of human perceptions of parallelity: we asked annotators on Amazon Mechanical Turk (MTurk)
to determine whether the image and slogan convey the same message
for a variety of ads. 
Then, we evaluated different strategies for their ability to identify ad image/slogan in parallel/non-parallel relationships on the collected dataset. We find that the proposed ensemble of cues outperforms standard image-text alignment methods.

The ability to determine whether an image and text are parallel without literally describing the same thing, has many applications. 
For example, advertising is a common marketing communication strategy that utilizes a visual form to deliver persuasive messages to its target audiences and sell a product, service or idea. 
A system that can understand the complex relationship between image and text is better equipped to understand the underlying message of the ad. Conversely, this system could also be used to help design more sophisticated ads.
Other applications include: flagging poor journalism by detecting unsubstantiated bias (e.g., recognizing that an unflattering image of a celebrity has no semantic association with the corresponding news story other than to bias the audience), generating creative captions (e.g., creating memes that deliver diverse messages vividly by accompanying different text with the same image) and promoting media literacy for educational purposes (e.g., teaching young consumers to recognize persuasive strategies employed by ads).

\vspace{-1em}
\section{Related Work}
\label{sec:related}
\vspace{-0.5em}

Prior work in matching images and text is primarily applied for image captioning or learning visual-semantic spaces. We show in experiments that this is not sufficient for understanding the relationship between persuasive images and persuasive text. A more sophisticated strategy is to determine how logically coherent an image and text pair is. This is similar to modeling {\em discourse relations} in natural language processing, which aims to identify the logical connection between two utterances in a discourse. 
Image-text coherence does not yet have the theoretical foundation or resource support that textual coherence has. 
We also discuss tasks related to judging image memorability and text abstractness, which we show have some relation to image-text parallelity (e.g. if the text is too abstract, it may not be able to convey the full message without help from the image). Finally, we describe prior work on understanding image advertisements and visual persuasion.

\vspace{-1em}
\paragraph{Visual-semantic embeddings and image captioning.}

Work in learning joint image-text spaces \cite{kiros2014unifying,Eisenschtat_2017_CVPR,Chen_2017_CVPR,Cao_2017_CVPR,Gomez_2017_CVPR, songfinebmvc17} aims to project an image and a matching piece of text in such a way that the distance between related text and image is minimized. 
For example, \cite{kiros2014unifying} use triplet loss where an image and its corresponding human-provided caption should be closer in the learned embedding space than image/caption that do not match.
\cite{Eisenschtat_2017_CVPR} propose a bi-directional network to maximize correlation between matching images and text.
None of these consider images with implicit or explicit persuasive intent, as we do. 
Image and video captioning \cite{vinyals2015show,Karpathy_2015_CVPR,Donahue_2015_CVPR,Johnson_2016_CVPR,Hendricks_2016_CVPR,Venugopalan_2017_CVPR,Vedantam_2017_CVPR,Gan_2017_CVPR,show_adapt_tell,Shetty_2017_ICCV,Krishna_2017_ICCV} is the task of generating a textual description of an image, or similarly, finding the textual description that best matches the image. 
These have been successful when depicting observable objects and their spatial relationships, but in our project, the interaction between visual and linguistic information can be hidden from the surface, which is more difficult to detect and requires novel techniques. 

\vspace{-1em}
\paragraph{Discourse relations.}
Linguistic units in text are connected logically, forming {\em discourse relationships} with each other. 
For example, a later sentence might \emph{elaborate} on an earlier sentence, it might serve as a \emph{contrast}, or it might state the \emph{result} of the earlier one. 
In this light, our problem of determining the relationship between image and text bears some resemblance to the problem of automatically determining discourse relations between sentences, for which there is extensive prior work
\cite{Prasad08thepenn,Pitler:2009:ASP:1690219.1690241,Park:2012:IID:2392800.2392818,biran-mckeown:2013:Short,qin-zhang-zhao:2016:EMNLP2016,chen-EtAl:2016:P16-13}. However, unlike the text-only case, we can neither rely on explicit linguistic cues nor on annotated resources such as the Penn Discourse Treebank.

\vspace{-1em}
\paragraph{Concreteness and specificity of text.}
Concrete words are grounded in things that we can sense (e.g., birds, flying, perfume) while abstract words refer to ideas that we do not physically perceive (e.g., cause, ``think different,'' tradition). 
A related concept is {\em specificity}, the level of details of the text being communicated (e.g. ``robin'' is more specific than ``bird'').
Resources and methods exist to predict concreteness and specificity 
\cite{coltheart1981mrc,turney-EtAl:2011:EMNLP,Li:2015:FAP:2886521.2886638}.
These textual stylistic clues may help to assess whether the text and image match in their specificity and concreteness.    

\vspace{-1em}
\paragraph{Image creativity and memorability.}

Images in the media are often more creative than typical photographs.
Several works examine the aesthetics of photos \cite{murray2012ava,schifanella2015image} and the creativity of videos \cite{redi20146}.
Some study the memorability of images \cite{khosla2015understanding,isola2014makes}. In our work, we consider memorability as a cue for the relationship between image and text; memorable images might be more creative (such as image C in Fig.~\ref{fig:concept}) and might indicate a ``twist'' and \textit{non-parallel} relationship between the image and text. 


\paragraph{Visual rhetoric and advertisements.}

There has \emph{not} been extensive work in visual rhetoric, which is especially common in advertisements. 
In a first work on visual rhetoric, 
\cite{joo2014visual} devised a method to predict which of two photos of a politician portrays that person in a more positive light, detecting implications that the photo conveys beyond the surface.
\cite{Hussain_2017_CVPR,ye2018advise} take another step towards understanding visual rhetoric, focusing on image and video advertisements. 
They formulate ad-decoding as answering a question about the reason why the audience should take the action that the ad suggests, e.g. ``I should not be a bully \emph{because words hurt as much as fists do}.''
However, their prediction method ignores any text in the advertisement images, which is crucial for decoding the rhetoric of ads.

\vspace{-1em}
\section{Dataset}
\label{sec:dataset}
\vspace{-0.5em}

We use images from the dataset of \cite{Hussain_2017_CVPR}, and annotate them on MTurk for our task. 
First, we randomly sample 1000 ads images that were predicted to be modern and not too wordy. We request a transcription of the important text (e.g. slogans) in the image, and labels regarding the relationship of the transcribed text and the image. 
We will make the dataset available for download upon publication.

\vspace{-1em}
\paragraph{Image selection.}

We want to determine how a concise piece of text interacts with the image. In order to focus our task on interesting relationships, we opted to filter images in two ways. 
First, we noticed that vintage ads often did not seem very creative in their use of language, compared to contemporary ads. Often the text was fairly straightforward and literal, while modern ads make clever use of idioms and wordplay, and aim to be concise and ``punchy''.
We observed that wordy ads were oftentimes also not creative. 
Thus, we implemented automatic methods to filter vintage or wordy ads. We trained a classifier to distinguish between 2080 vintage ads crawled from the Vintage Ad Browser\footnote{http://www.vintageadbrowser.com} and 4530 modern ads from Ads of the World\footnote{https://www.adsoftheworld.com}. We used the ResNet-50 architecture \cite{he2016deep} pre-trained on ImageNet \cite{russakovsky2015imagenet} and further trained on these vintage/modern ads, achieving 0.911 precision and 0.950 recall on a held-out validation set.
We applied it to images from the dataset of \cite{Hussain_2017_CVPR}, and removed 11582 images that were predicted to be vintage. 
We also used optical character recognition (OCR) from the Google Cloud Vision API\footnote{https://cloud.google.com/vision/} and counted the length of the automatically extracted text fragments in each image. We removed 33503 images containing less than 10 or more than 80 words. 
In the end, we were left with 19747 images, from which we sampled 1000 for human annotation.

\vspace{-1em}
\paragraph{Transcription.}

Automatic text recognition did not work sufficiently well, so we asked humans to manually transcribe the persuasive text in the ads images. 
We instructed workers to ignore the following: logos, legalese, date and time, long paragraph of description, urls, hashtags and non-English characters.
We emphasized they should transcribe that text which carries the core message of the image.

\vspace{-1em}
\paragraph{Relationship annotation.}

\begin{figure}[t]
\includegraphics[width=1\linewidth]{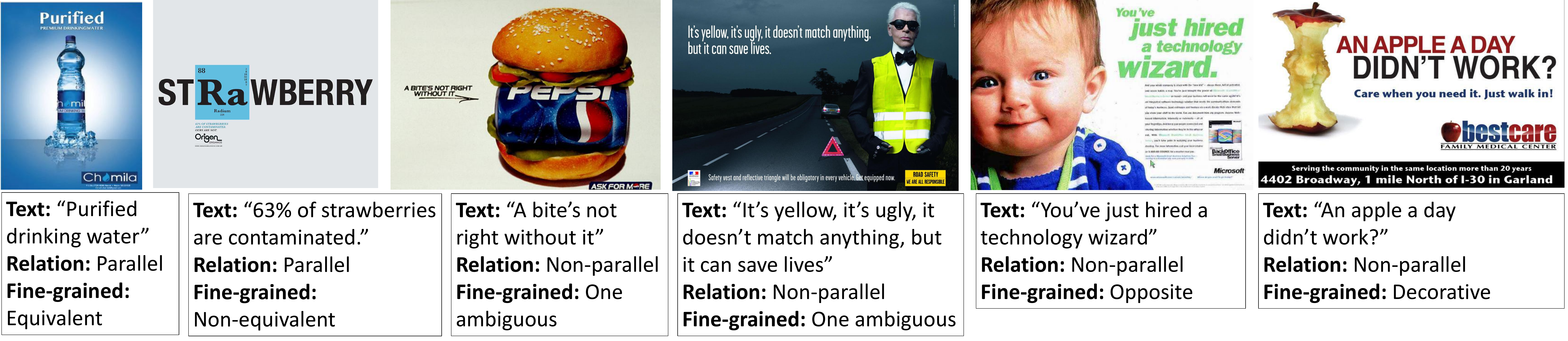}
\caption{Image-text pairs and their relationship (parallel/non-parallel) with rationales. Note: In the second image, Ra is the chemical element Radium.}
\vspace{-1em}
\label{fig:examples}
\end{figure}

For each image, the key question we want to answer is: Are the text and image \emph{parallel}? 
We define \emph{parallel} as 
image and text individually aiming to convey the same message as the other; see the first two examples in Fig.~\ref{fig:examples}. 
We instruct annotators that: ``If you have no difficulty understanding the ad with only the highlighted text, and you have no difficulty understanding it with only the image, the relationship is \emph{parallel}.'' 
In contrast, for \emph{non-parallel} pairs, ``one of text or image might be fairly unclear or ambiguous without the other, text and image alone might imply opposite ideas than when used in combination, or one might be used to attract the viewer's attention while providing no useful information.''

\vspace{-1em}
\paragraph{Additional questions.}

We also ask our annotators two additional questions. First, we ask them to describe the message that they perceive in the ad: ``Could you figure out the message of this ad? Please give an explicit statement in one or two sentences.''  We found that asking annotators to type the message of the ad helped them to think more thoroughly about the ad image, which resulted in higher-quality results. We are not interested in this annotation beyond quality control; further, \cite{Hussain_2017_CVPR}'s dataset already provides this info.

\vspace{-1em}
\paragraph{Fine-grained relationships.}

Second, we asked annotators to provide some rationale for their answer in the main parallel/non-parallel question. Annotators could choose from the following five rationale options (or type their own):
\begin{enumerate}[noitemsep,nolistsep]
{\small
\item \textbf{Equivalent parallel:} The text and image are completely equivalent: They make exactly the same point, equal in strength (e.g. the first image in Fig.~\ref{fig:examples}).
\item \textbf{Non-equivalent parallel:} They try to say the same thing, but one is more detailed than the other: They express same ideas but at different level (e.g. the second image in Fig.~\ref{fig:examples}).
\item \textbf{One ambiguous (non-parallel):} Text or image is fairly unclear or ambiguous without the other.
\item \textbf{Opposite (non-parallel):} Text and image alone imply opposite ideas than when used together.
\item \textbf{Decorative (non-parallel):} The main idea of the ad is conveyed by just the text (or image) alone; the other is primarily decorative.
}
\end{enumerate}

The first two rationales are meant to accompany a \textit{parallel} response, and the next three a \textit{non-parallel} one.
The boldface fine-grained relationship label is internal, and not provided to annotators. 
We designed this question for improved annotation quality, as we found that in our pilot collection, annotators chose the \textit{parallel} option too often due to carelessness.
By making them think about the ad harder, we obtained higher-quality responses. 

\vspace{-1em}
\paragraph{Quality control.} 

We only allow MTurk workers who have completed over 1000 HITs and have 95\%+ approval rate to participate in our study. For each image/text pair, we collected five relationship labels, and performed a majority vote to obtain the final ground-truth. 
To obtain high-quality labels, we launched the annotation task in batches (each containing 20 image/text pairs),
and embedded two quality control questions in each batch. One is a question for which we know the relation between text and image in advance, and the answer is unambiguous. The other is a repeat; we expect the annotators to answer the same way both times if they were paying attention. We removed data from annotators that failed both of these tests, or wrote clearly careless answers to our question about the ad's message.

\vspace{-1em}
\paragraph{Ambiguity.} 
One characteristic we observed from collecting such subjective annotations is the lower agreement across workers compared with other objective annotation tasks. For example, for the parallelity annotation task, nearly 40\% of all collected annotations have a 3/2 or 2/3 split between positive and negative choices. After manual inspection we noticed that the difference might come from the fact that image and text might be parallel \textit{on different levels}. Although this interesting trend deserves further exploration, in this study we removed these ambiguous cases and only focused on the instances with high inter-rater agreement.

\vspace{-1em}
\section{Approach}
\label{sec:approach}
\vspace{-0.5em}

We develop an ensemble of predictors for the relationship between image and text. Each captures a different aspect of parallelity, and when all are combined, they predict the relationship much more reliably than prior methods for image captioning. 
Note that even though some classifiers use information only from a single channel (text or image), they still demonstrate competitive predictive power for modeling the relationship. One possible explanation is they capture how ads are created by designers. For example, if wordplay is used in the slogan it is usually a strong indicator of ambiguity and non-parallelity.

\vspace{-1em}
\subsection{Base features}
\label{sec:base}

We develop the following features to recognize parallel relationships, then train classifiers using them. These capture a variety of aspects, including alignment and distances between channels, as well as stylistic and rhetorical nuances of individual channels.

\vspace{-1em}
\paragraph{1: Visual Semantic Embedding (VSE).}

Deep neural networks trained on massive datasets have demonstrated their potential to capture semantic meaning in many tasks. 
To modify existing visual-semantic embeddings for our task,  
we apply the pre-trained VSE model from \cite{kiros2014unifying} on our dataset, converting ad images and the accompanying slogans into fixed-length embeddings which are concatenated as a feature set. 
VSE is expected to be good at detecting literal parallel cases (i.e. \textit{equivalent parallel}).
 
\vspace{-1em}
\paragraph{2: Concepts alignment.}

While VSE offers a means to compare the image and text in their entirety, it may also be useful to determine whether a certain concept is explicitly mentioned in both the image and the text. As a heuristic, concept alignment may give us a high-precision classifier for identifying parallel cases (and as a corollary, a high-recall classifier for identifying non-parallel cases). 
We use the general-purpose recognition API provided by Clarifai\footnote{https://clarifai.com/developer/guide/} to extract dominating concepts from ad images. Typically each image will be tagged with 20 relevant concepts together with probabilities, and the concepts cover a wide range of objects, activities and scenes. For text, we preprocess the transcribed slogans using the NLTK package \cite{loper2002nltk} to perform automatic part-of-speech tagging. After removing stop-words, we empirically select \texttt{NOUN}, \texttt{ADJ}, \texttt{ADV} and 
\texttt{VERB} to be key concepts in text. 
We then compute pairwise semantic distances between the collected image concepts and text concepts. We use a pre-trained word2vec \cite{mikolov2013distributed} model as the bridge to convert concepts into a semantically meaningful vector representation, then measure their cosine distances.

\vspace{-1em}
\paragraph{3: Agreement from single channel prediction.}
If the viewer can deduce the correct message from either channel alone, then using either text or image would be sufficient for advertising, and the relationship might be \textit{parallel}, but if the meaning is unclear with one channel disabled, then both channels are indispensable, and the relationship might be \textit{non-parallel}. 
To capture this intuition, the most fitting task is to evaluate whether the intended message can be perceived with information from a single channel, but this task is very challenging as shown in \cite{Hussain_2017_CVPR}. Here we choose the prediction of the ad's topic as a proxy task. 

For images, we 
infer the ad topic using a fine-tuned ResNet \cite{he2016deep} model, and take the predicted probability distribution 
as the information provided from the image channel. 
For text, 
since we have only a limited number of transcribed slogans, we use the Google Cloud Vision API for OCR and extract text for the full ads dataset from \cite{Hussain_2017_CVPR}. 
After removing non-English words and stop-words, we group the words in the extracted text according to their ad topic annotation and perform TF-IDF analysis. The resultant adjusted counts represent the relative frequency for individual words to appear in various topics. For inference, we let individual words vote for the topics using weights proportional to the TF-IDF scores. 
We apply softmax to convert the votes into a probabilistic distribution over topics, and the two predicted topic distributions are then concatenated as the feature.

\vspace{-1em}
\paragraph{4: Surprise factor.}

In addition to the contrast across image and text channels, empirical evaluations indicate that the relationship between concepts within a single channel is informative as well. For example, using words that contradict each other in the slogan is sufficient to make an impression (``less is more'') by itself, and an image with objects that do not normally co-occur (like cats in a car ad) catches the viewer's attention. 
We calculate the pairwise distances between concepts extracted within the image and text channels, respectively, and use the normalized histogram of the distribution of cosine distances as a feature vector.

\vspace{-1em}
\paragraph{5: Image memorability.}

Studies have shown that memorable and forgettable images have different intrinsic visual features. For ads, memorability is an especially interesting property as one of the goals for ads is to be remembered.  Therefore a reasonable guess is that visual memorability can affect how designers arrange elements within ads, including both image and text. Here we adopt the model in \cite{khosla2015understanding} and perform inference on our collected ads, and the 1D resultant memorability score (higher means more memorable) is used as a feature.

\vspace{-1em}
\paragraph{6: Low-level vision features.}

Low-level vision features like HoG \cite{dalal2005histograms}, which capture gradient orientation statistics, have been widely used for image recognition and object detection.
To obtain a fixed-length representation for ads with varied sizes, we resize all the images to be $128\times128$ and use $9$ orientations, $16\times16$ pixels per cell and $2\times2$ cells per block. 

\vspace{-1em}
\paragraph{7: Slogan specificity and concreteness.}

Sentence specificity and word concreteness have been recognized as important characterizations of texts \cite{paivio1971imagery} for language tasks such as document summarization and sentiment analysis. Here we adopt the model from \cite{Li:2015:FAP:2886521.2886638} to predict a specificity score for the slogans we collected from manual transcription. 
We transformed the transcriptions into four different forms (original, all uppercase, all lowercase and only first character of sentence is capitalized), and the specificity scores for all these four forms are concatenated. Similarly, we extract the concreteness score for individual words within slogans using the MRC dataset \cite{coltheart1981mrc}, and calculate the max, median, mean and stdev.

\vspace{-1em}
\paragraph{8: Lexical ambiguity and polysemy.}
Polysemy is a prevailing phenomenon in languages, and this lexical ambiguity has been widely used in advertising because the varying interpretation makes the viewer more involved in understanding the ads, and subsequently makes the ad more impressive. We use WordNet \cite{miller1995wordnet} to query each word within the slogan, and collect statistics on the number of plausible meanings 
to estimate the ambiguity in the slogan.

\vspace{-1em}
\paragraph{9: Topics.}

We use a one-hot vector for the ground-truth topic annotation of ads \cite{Hussain_2017_CVPR}, following an intuitive clue that 
ads belonging to the same theme could use similar strategies for image and text composition.

\vspace{-1em}
\subsection{Combining clues from different classifiers}
\label{sec:combined}

We attempted several ways to combine our cues from the previous section.
One can concatenate all previously described features, perform a majority vote over the individual classifier responses, or learn a weighted voting scheme via bagging and boosting.
While several of these gave promising results, the best approach was the following.
Rather than constructing one classifier towards the overall goal, we first focus on how each individual classifier performs, and then use auxiliary features to help choose reliable classifiers under different scenarios. Specifically, since our base classifiers capture different perspectives of parallelity, they demonstrate varied performance on different samples. Intuitively, if 
an auxiliary classifier can capture contextual information to judge which classifier to trust for a specific test instance, performance should improve compared to weighing all classifiers equally. 


Consider one sample for illustration so we can omit the sample subscript. We construct $m$ sets of features (corresponding to the subsections in Sec.~\ref{sec:base}); let $X_i$ be the $i$-th set. We first train $m$ SVM classifiers $f_i(X_i)$, one per feature set.
Then we take the output $\hat{y}_i = f_i(X_i)$ and its probability estimate $\hat{p}_i$.
We further construct a new set of labels, $o_i = \mathbbm{1}_{\{y = \hat{y}_i\}}$, and train $g_i(\hat{p}_1, ..., \hat{p}_m)$ for each $o_i$,
to obtain prediction $\hat{o}_i = g_i(\hat{p}_1, ..., \hat{p}_m)$, measuring whether each base prediction is reliable for certain contexts.
Finally, we take the $k$ largest $\hat{o}_i$, and use the corresponding $\hat{y}_i$ to get a majority vote (we choose $k=5$ in our experiments).

\vspace{-1em}
\section{Experimental Validation}
\label{sec:results}
\vspace{-0.5em}

In this section, we verify that our proposed features can help distinguish between \textit{parallel} and \textit{non-parallel} ads better than standard methods (e.g. VSE) can, especially for the harder cases (such as distinguishing between \textit{non-equivalent parallel} ads and \textit{non-parallel} ads).

\vspace{-1em}
\subsection{Experimental setup}
\begin{table}
\begin{center}
\begin{tabular}{|l|c|c|c|}
\hline
& Counts & Parallel Score & Equivalent Score \\
\hline\hline
Non-parallel & 169 & 0.291 (0.141) & 0.143 (0.148) \\
\hline
Parallel Equivalent & 84 & 0.924 (0.097) & 0.767 (0.114) \\
Parallel Non-equivalent & 84 & 0.857 (0.090) & 0.210 (0.127)\\
\hline
\end{tabular}
\end{center}
\caption{Statistics on the dataset constructed for our experiments. The parallel score mean (stdev) come from the main annotation, and the equivalent score from the fine-grained rationale. We analyze performance on the non-equivalent examples as they are more challenging.}
\vspace{-1em}
\label{table:statistics}
\end{table}

For each image-text pair in our collected dataset, we calculate the \textit{parallel score} according to the responses from all annotators by the percentage of \textit{parallel} responses (e.g. 0.6 if 3 out of 5 annotators choose \textit{parallel}).
To validate our proposed strategy, we remove some ambiguous cases and construct a balanced dataset. We choose those image/text pairs for which more than half of annotators agree to be \textit{non-parallel} as the negative samples, and sample roughly the same number of positive samples from ads that have \textit{parallel} score higher than 0.8 (over four-fifths agreement on \textit{parallel}). Since \textit{parallel} ads are varied, we use the fine-grained rationales (Sec.~\ref{sec:dataset}) to construct a balanced set within \textit{parallel.} 
We make sure that \textit{equivalent} and \textit{non-equivalent} ads (corresponding to the first and second rationales) are balanced within the \textit{parallel} category. 
In Table \ref{table:statistics} we report statistics about the constructed dataset, which clearly show the distinct rating scores between \textit{parallel} and \textit{non-parallel} ads, as well as \textit{equivalent} and \textit{non-equivalent} ads.
Note that our main task is to predict \textit{parallel} while the equivalent/non-equivalent task is only used for further analysis.

\vspace{-1em}
\subsection{Results}



\begin{table}[t]
\begin{center}
\begin{tabular}{|c|c||c||c|c|c|}
\hline
\multirow{3}{*}{Feature Set} & \multirow{3}{*}{Dim} & \multirow{3}{*}{Avg accuracy} & \multicolumn{3}{c|}{Per-class avg accuracy} \\
\cline{4-6}
& & & \multirow{2}{*}{Non-parallel} & \multicolumn{2}{c|}{Parallel} \\
\cline{5-6}
& & & & Equivalent & Non-equiv \\
\hline\hline
1: VSE & 2048 & 0.602 & 0.574 & \emph{0.667} & 0.595 \\
2: Alignment & 85 & 0.581 & 0.598 & \emph{0.595} & 0.536 \\
3: Topic Agreement & 80 & 0.516 & 0.728 & 0.298 & \emph{0.310} \\
4a: Image Surprise & 65 & 0.549 & 0.550 & \emph{0.595} & 0.500 \\
4b: Text Surprise & 70 & 0.596 & 0.598 & 0.548 & \emph{0.643} \\
5: Memorability & 1 & 0.567 & 0.408 & \emph{0.774} & 0.679 \\
6: HoG & 1764 & 0.582 & 0.728 & 0.405 & \emph{0.464} \\
7a: Specificity & 4 & 0.534 & 0.491 & 0.524 & \emph{0.631} \\
7b: Concreteness & 4 & 0.525 & 0.361 & 0.679 & \emph{0.702} \\
8: Polysemy & 4 & 0.504 & 0.308 & 0.643 & \emph{0.762} \\
9: Topic & 40 & 0.539 & 0.402 & \emph{0.702} & 0.655 \\
\hline\hline
Combination & 11 & \textbf{0.655} & 0.633 & 0.702 & 0.655 \\
\hline
\end{tabular}
\end{center}
\caption{Experimental results using individual classifiers as well as our combined classifier, on predicting \textit{parallel} vs \textit{non-parallel}. 
Accuracy is whether the prediction matches the ground truth, and we measure it over the entire dataset and over three fine-grained subcategories.
The method performing best overall is in \textbf{bold}. For the last two columns, we \emph{italicize} whether an individual feature does better on the harder case
of correctly classifying non-equivalent parallel ads, 
or the easier case of equivalent parallel. 
Our combined method outperforms the VSE baseline, and our features generally do better on the harder case.}
\vspace{-1em}
\label{table:result}
\end{table}

We show our quantitative results in Table \ref{table:result}, including performances of individual classifiers from Sec.~\ref{sec:base} and the combination of features (Sec.~\ref{sec:combined}). For fair comparison we use SVMs with linear kernels for all feature sets, and report performance measured from five-fold cross validation. 
We observe the five individual methods that are most competitive in terms of overall average (third column) are VSE, Text Surprise, HoG, Alignment and Memorability. However, the combined classifier performs best, and outperforms VSE by 9\%.

We are particularly interested in the type of samples that each method classifies correctly. The distinction between \textit{parallel equivalent} (such as the first image in Fig.~\ref{fig:examples}) and \textit{non-parallel} is much more obvious, than the distinction between \textit{parallel non-equivalent} (such as the second image in Fig.~\ref{fig:examples}) and \textit{non-parallel}. 
Thus, we hope our methods which were specifically designed to handle less obvious cases can do well on the harder cases. 
This is indeed true for most of our methods, especially Text Surprise, HoG, Specificity, Concreteness and Polysemy. 
Topic Agreement does not help much for overall accuracy but it performs best for capturing \textit{non-parallel} image-text pairs, tied with HoG. 
Memorability is the best across all methods for distinguishing between \textit{parallel equivalent} and \textit{non-parallel} ads.

\begin{figure}[t!]
\centering
\includegraphics[width=\linewidth]{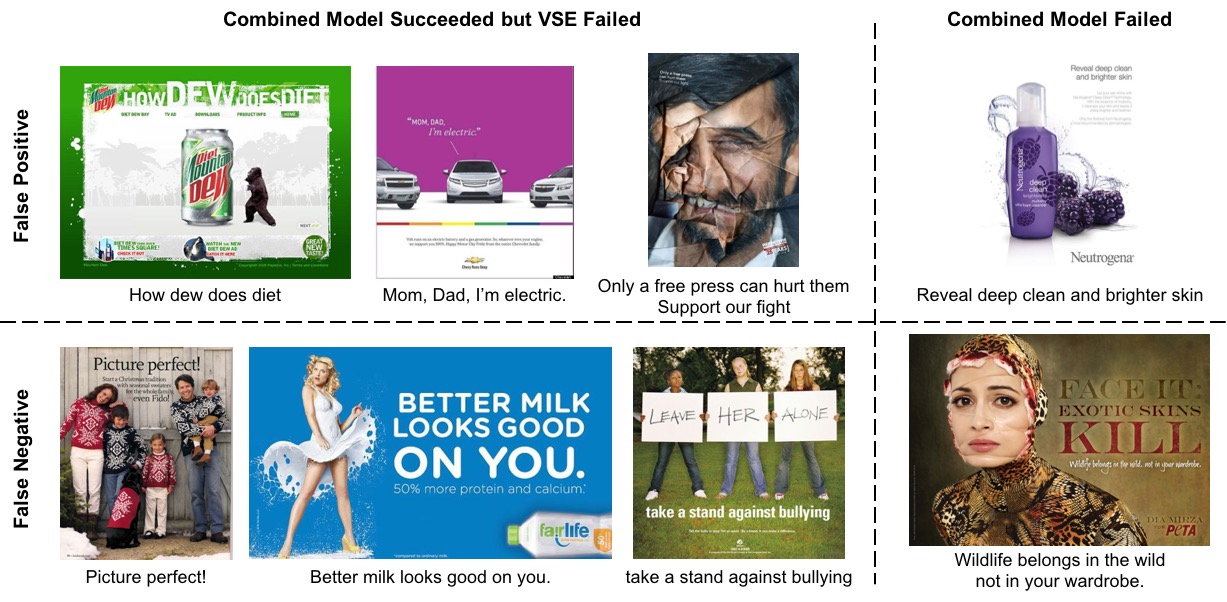}
\vspace{-0.75cm}
\caption{Qualitative examples from our combined model and VSE. (Top) False positive: Non-parallel cases classified as parallel by VSE or our model. (Bottom) False negative: Parallel cases classified as non-parallel. (Left): Our combined model correctly captures the image/text relationship but VSE fails. (Right) Both our combined model and VSE fail.}
\vspace{-1em}
\label{fig:qualitative}
\end{figure}

Finally, we give some qualitative examples in Fig.~\ref{fig:qualitative}, comparing our model to the second-best method, VSE. 
One example in which our model outperforms VSE is the electric car ad. Here, VSE mistakenly predicts \textit{parallel}, but the difference between ``Mom'', ``Dad'' and ``electricity'' is captured by Text Surprise. Additionally, HoG and Topic Agreement also predict "non-parallel." While some individual classifiers choose \textit{parallel}, our combination model has correctly relied on Text Surprise, HoG, and Topic Agreement to make the right prediction. Another success is the ``Picture perfect!'' example. The detected image concepts are all semantically close (child, family, people, etc.) and such trends are captured by Image Surprise for \textit{parallel.} Our combination model picks it as a trustworthy classifier, along with Alignment and Topic, and makes the correct prediction.

Our model makes mistakes, especially for semantically subtle ads. In the ``Reveal deep clean and brighter skin'' example, the text highlights the \textit{effect} after using the makeup, while image illustrates the makeup itself. Human raters have no problem recognizing the differences, while VSE, Topic Agreement (both image and text are indicative of makeup), Image Surprise (common and natural image, little surprise factor) and Concreteness (very concrete expression) all predict \textit{parallel.} Although Alignment predicts \textit{non-parallel} (image and text do not align well), our combination model incorrectly goes with the majority prediction.

\vspace{-1em}
\section{Conclusion}
\label{sec:conclusion}
\vspace{-0.5em}

We proposed the novel problem of analyzing non-literal relationships between persuasive images and text. We developed features that aim to approximate some of the persuasive strategies used by ad designers.  
We tested these features on the task of predicting whether an image and text are \textit{parallel} and convey the same message as each other without saying the exact same thing, for which we crowdsourced high-quality annotations. We showed that our combined classifier, which uses our proposed features, significantly outperforms a baseline method developed for literal image-text matching, as in image captioning. 

\vspace{-1em}
\paragraph{Acknowledgements.}
We would like to thank Daniel Zheng for helping in analyzing the sentence concreteness, and Keren Ye and Xiaozhong Zhang for valuable discussions throughout the project. We would also like to thank the anonymous reviewers for their helpful comments.
This material is based upon work supported by the National Science Foundation under Grant
Number 1718262 and an NVIDIA hardware grant. Any opinions, findings, and conclusions or recommendations expressed in this material are those of the author(s) and do not necessarily reflect the views of the National Science Foundation. 

\bibliography{refs_main,refs_persuasion,references_ak,references_ms,references_rh}
\end{document}